\documentclass[11pt,a4paper]{article}
\usepackage[hyperref]{emnlp2020}
\usepackage{times}
\usepackage{latexsym}

\usepackage{microtype}

\aclfinalcopy 

\usepackage{url}
\usepackage{soul}
\usepackage{graphicx}
\usepackage[toc,page]{appendix}
\usepackage[export]{adjustbox}
\usepackage{amsmath, amsfonts}
\usepackage{graphicx}
\usepackage{pifont}
\usepackage{booktabs}
\usepackage{multirow}

\usepackage{rotating}
\usepackage{subfig}
\title{Counterfactually-Augmented SNLI Training Data Does Not Yield Better Generalization Than Unaugmented Data}

\author{
William Huang \\
New York University \\
{\tt will.huang@nyu.edu} \\ \And
Haokun Liu \\
New York University \\
{\tt haokunliu@nyu.edu}\\ \And
Samuel R. Bowman \\
New York University \\
{\tt bowman@nyu.edu}}

\begin{document}

\maketitle
\begin{abstract}
A growing body of work shows that models exploit annotation artifacts to achieve state-of-the-art performance on standard crowdsourced benchmarks---datasets collected from crowdworkers to create an evaluation task---while still failing on out-of-domain examples for the same task. Recent work has explored the use of counterfactually-augmented data---data built by minimally editing a set of seed examples to yield counterfactual labels---to augment training data associated with these benchmarks and build more robust classifiers that generalize better. However, \citet{khashabi2020pertboolq} find that this type of augmentation yields little benefit on reading comprehension tasks when controlling for dataset size and cost of collection. We build upon this work by using English natural language inference data to test model generalization and robustness and find that models trained on a counterfactually-augmented SNLI dataset do not generalize better than unaugmented datasets of similar size and that counterfactual augmentation can hurt performance, yielding models that are less robust to challenge examples. Counterfactual augmentation of natural language understanding data through standard crowdsourcing techniques does not appear to be an effective way of collecting training data and further innovation is required to make this general line of work viable.
\end{abstract}

\section{Introduction}
\label{sec:intro}
While standard crowdsourced benchmarks have helped create significant progress within natural language processing (NLP), a growing body of evidence shows the existence of exploitable annotation artifacts in these datasets \citep{gururangan2018artifactsNLI, poliak2018hypothesisonly, tsuchiya18RTEhypothesis} and that models can use artifacts to achieve state-of-the-art performance on these benchmarks \citep{ mccoy2019HANS, naik2018NLIstress}. The existence of these artifacts makes it difficult to predict out-of-domain generalization and creates uncertainty around the abilities these tasks are designed to test. 

Recent work has explored using counterfactually-augmented datasets to address annotation artifacts with the intent to build more robust classifiers \citep{kaushik2020cnli, khashabi2020pertboolq}. These datasets are collected by first sampling a set of seed examples and then creating new examples by minimally editing the seed examples to yield counterfactual labels. This type of data collection has been found to mitigate the presence of artifacts in SNLI \citep{bowman2015SNLI} and is presented as a way to ``elucidate the difference that makes a difference'' \citep{kaushik2020cnli}. Further, \citet{khashabi2020pertboolq} present this as an efficient method to collect training data yielding models that are ``more robust to minor variations and generalize better'' \citep{khashabi2020pertboolq}. However, they also find that unaugmented datasets yield better performance than datasets with 50-50 original-to-augmented data when controlling for training set size and annotation cost. 

In our work, we further study whether training with counterfactually-augmented data collected through standard crowdsourcing methods yields models with better generalization and robustness by focusing on the domain of natural language inference (NLI): the task of inferring whether a \textit{hypothesis} is true given a true \textit{premise}. We train and compare RoBERTa \citep{liu2019RoBERTa} trained on three different datasets: (1) the counterfactually-augmented natural language inference (CNLI) training set of 8.3k seed and augmented SNLI examples from \citet{kaushik2020cnli}, (2) a subsampled set of 8.3k unaugmented SNLI examples to control for size, and (3) the 1.7k CNLI seed examples originally sampled from SNLI. We then compare model performances on MNLI \citep{williams2018MNLI}---a dataset for the same task with examples out-of-domain to SNLI---and two diagnostic sets \citep{naik2018NLIstress, wang2019GLUE}.

We find that RoBERTa trained on CNLI yields similar performance on out-of-domain MNLI examples when compared to the unaugmented subsampled SNLI training set and that including counterfactually-augmented examples to the CNLI seed set improves generalization. Further, we find that the improvement over seed examples correspond to an increase in n-grams from the addition of augmented examples, roughly doubling the number of 4-grams, and may be a result of improved lexical diversity from a larger training set. While we see similar trends in most of our diagnostic evaluations, we also find evidence that including augmented examples can yield worse performance than only training with seed examples.

While there is evidence of the benefits of using this type of data for model evaluation \citep{gardner2020augmenteddata}, we find that using counterfactually-augmented data for training yields \textit{less} robust models. We argue that further innovation is required to effectively crowdsource counterfactually-augmented natural language understanding (NLU) data for training more robust models with better generalization.

\section{Related Work}
\label{sec:related}
Recent works show that several NLI benchmark datasets contain exploitable annotation artifacts. Several studies \citep{poliak2018hypothesisonly, gururangan2018artifactsNLI, tsuchiya18RTEhypothesis} show that models trained on hypothesis-only examples manage to perform as much as 35 points higher than chance. \citet{gururangan2018artifactsNLI} also find negation words such as \textit{no} or \textit{never} are strongly associated with \textit{contradiction} predictions. Other works \citep{naik2018NLIstress, mccoy2019HANS} find that models can exploit premise-hypothesis word overlap to achieve state-of-the-art performance on benchmarks by using associations of high overlap with \textit{entailment} predictions and low overlap with \textit{neutral} predictions.

\citet{nie2020adversarialNLI} use an adversarial human-and-model-in-the-loop procedure to address these concerns in Adversarial NLI (ANLI). Using a model in the loop makes ANLI inherently adversarial towards the model used, and we instead focus on naturally collected human-in-the-loop augmented data.

\citet{kaushik2020cnli} crowdsource counterfactually-augmented NLI examples that reduce the presence of hypothesis-only bias in SNLI by providing a set of seed examples to crowdworkers and prompting them to minimally edit either the hypothesis or premise to yield a counterfactual label. \citet{khashabi2020pertboolq} present this type of data collection as an efficient method to build training sets yielding robust models that generalize better by crowdsourcing counterfactually-augmented BoolQ examples. However, they also find that augmented datasets yield similar to worse performance when the cost of augmenting an example is no cheaper than collecting a new one and the datasets are controlled for size. We differ from \citet{kaushik2020cnli} by focusing on performance on out-of-domain examples and from \citet{khashabi2020pertboolq} by focusing on the task of NLI instead of reading comprehension.

\citet{gardner2020augmenteddata} use contrast sets written manually by NLP researchers to evaluate models on various annotated tasks. They show that most datasets require 1-3 minutes per augmented example, taking 17-50 hours to create 1,000 examples. We differ by using crowdsourced counterfactually-augmented data and focusing on their use for training instead of evaluation.

\section{Experimental Setup}
\label{sec:experiments}
We perform two experiments to study the effects of counterfactually-augmented NLI training data. All experiments use RoBERTa trained on SNLI, CNLI, or CNLI seed examples originally sampled from SNLI and compare performances on various tasks. We first compare MNLI performances to evaluate the impact on model generalization to out-of-domain data. We then use the diagnostic examples from \citet{naik2018NLIstress} and the GLUE diagnostic set \citep{wang2019GLUE} to study model robustness to challenge examples.

\paragraph{Training Data} In SNLI, \citet{bowman2015SNLI} prompt crowdworkers with a scene description premise to collect three hypothesis sentences corresponding to \textit{entailment}, \textit{neutral}, and \textit{contradiction} labels, yielding 570k English premise-hypothesis pairs. \citet{kaushik2020cnli} collect CNLI examples by prompting crowdworkers to minimally edit seed examples sampled from SNLI to yield counterfactual labels.

For our training data, we use a subsampled set of 8.3k examples of SNLI, the CNLI training set of 8.3k examples, and the 1.7k CNLI seed examples sampled from SNLI that is also included in the CNLI training set. We subsample SNLI to control for the fact that CNLI only consists of 8.3k examples. We subsample five sets of 8.3k SNLI examples and report results across these five.

\paragraph{Out-of-Domain Set}
We treat MNLI as our out-of-domain NLI evaluation data. In collecting MNLI examples, \citet{williams2018MNLI} follow a similar data collection framework while expanding the diversity of their premises by sourcing them from ten sources of freely available text, yielding 433k English premise-hypothesis pairs. The data set includes 393k training examples from five of the ten sources, 20k validation examples, and 20k test examples. The validation and test examples are split in half between \textit{matched} and \textit{mismatched} examples, where \textit{matched} examples come from the same five sources as training examples and \textit{mismatched} examples come from the remaining five sources. We report validation accuracy for the combined MNLI validation set.

\paragraph{Diagnostic Sets}
\citet{naik2018NLIstress} provide NLI diagnostic sets of automatically generated challenge examples based on MNLI. These sets are split into six categories named Antonymy, Numerical Reasoning, Word Overlap, Negation, Length Mismatch, and Spelling Error. As part of GLUE, \citet{wang2019GLUE} provide NLI diagnostic sets of challenge examples aimed to evaluate reasoning abilities related to four broad categories: Lexical Semantics, Predicate-Argument Structure, Logic, and Knowledge. We use these sets to test model robustness to challenge examples. We refer the reader to \citet{naik2018NLIstress} and \citet{wang2019GLUE} for additional details on each diagnostic set.

\citet{mccoy2019HANS} provide similar adversarial examples, but we find them too difficult for our models, with performance consistently below 3\%, so we do not report performance in detail. 

\paragraph{Implementation}
\label{subsec:implementation}
Our code\footnote{\url{https://github.com/nyu-mll/CNLI-generalization}} builds on \texttt{jiant v2 alpha} \citep{wang2019jiant}. All experiments use \texttt{roberta-base}. For each round of training, we perform 20 runs and randomly search the hyperparameter space of learning rate \{1e-5, 2e-5, 3e-5\}, batch size \{32, 64\}, and random seed. Given the small training set size and stability benefits from longer training found in \citet{mosbach2020stabilityBERT}, we train each run for 20 epochs using early stopping based on the respective validation sets.

\section{Results}
\label{sec:results}
\paragraph{Generalization to MNLI}
\begin{figure}[t]
    \centering
    \includegraphics[width=0.48\textwidth]{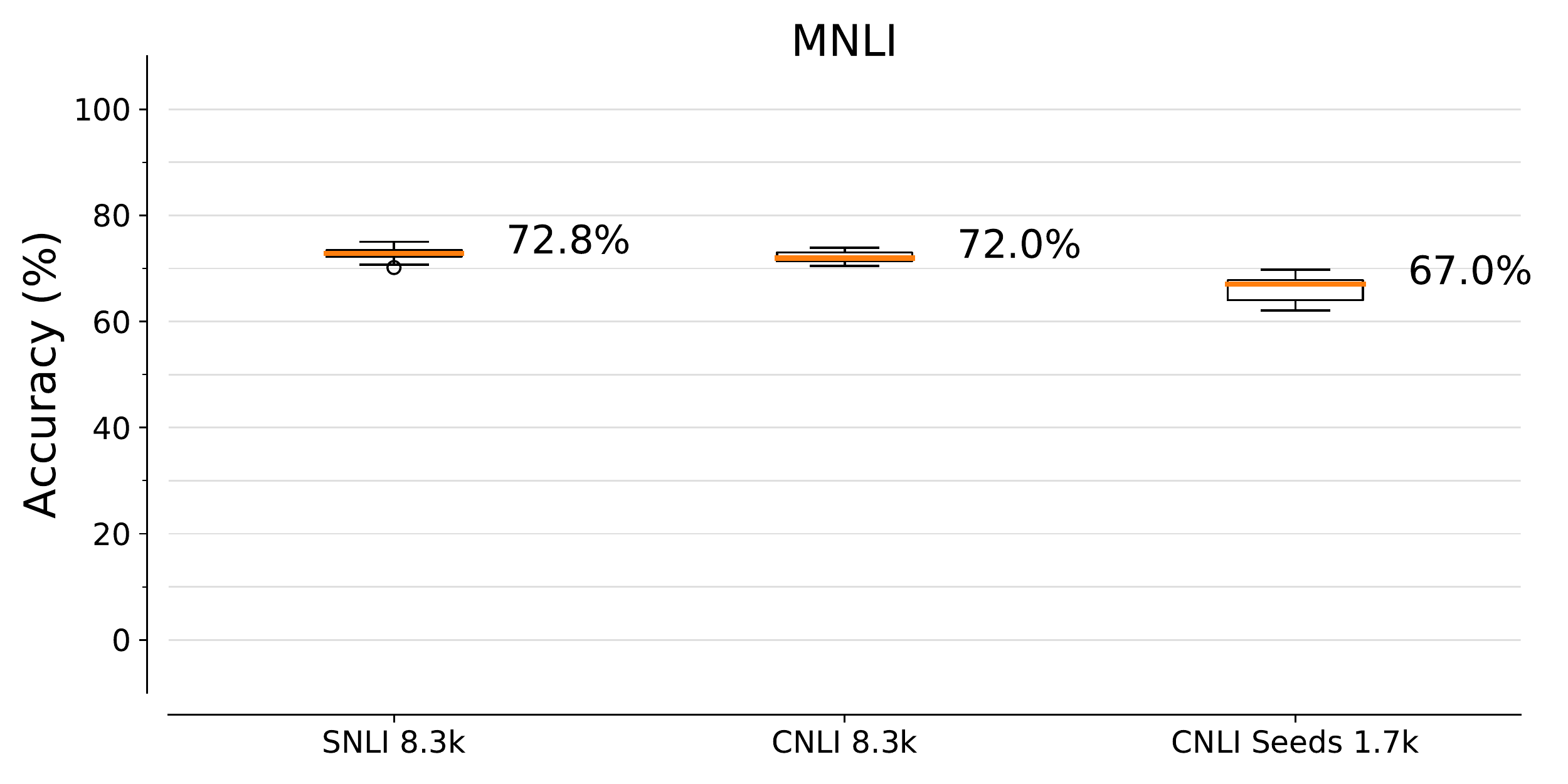}
    \caption{Combined MNLI \textit{matched} and \textit{mismatched} validation accuracy trained on subsampled SNLI, CNLI, and CNLI seed examples. The orange line and label indicate the median score.}
    \label{fig:mnli}
\end{figure}

From the median scores in Figure \ref{fig:mnli}, we see that models trained on CNLI perform no better than models trained on a comparably large sample of unaugmented SNLI examples. This is in line with findings from \citet{khashabi2020pertboolq}, where training with their minimally perturbed BoolQ dataset of seed and augmented examples yields similar or worse performance on out-of-domain tasks compared to the original BoolQ training set. Additionally, the improvement of CNLI over the 1.7k seed examples shows that counterfactual examples are somewhat helpful when they are strictly additive, as in \citet{khashabi2020pertboolq}.

\paragraph{Robustness to Diagnostic Sets}
\begin{figure*}[t]
    \centering
    \subfloat[a][]{
            \includegraphics[width=0.8\textwidth]{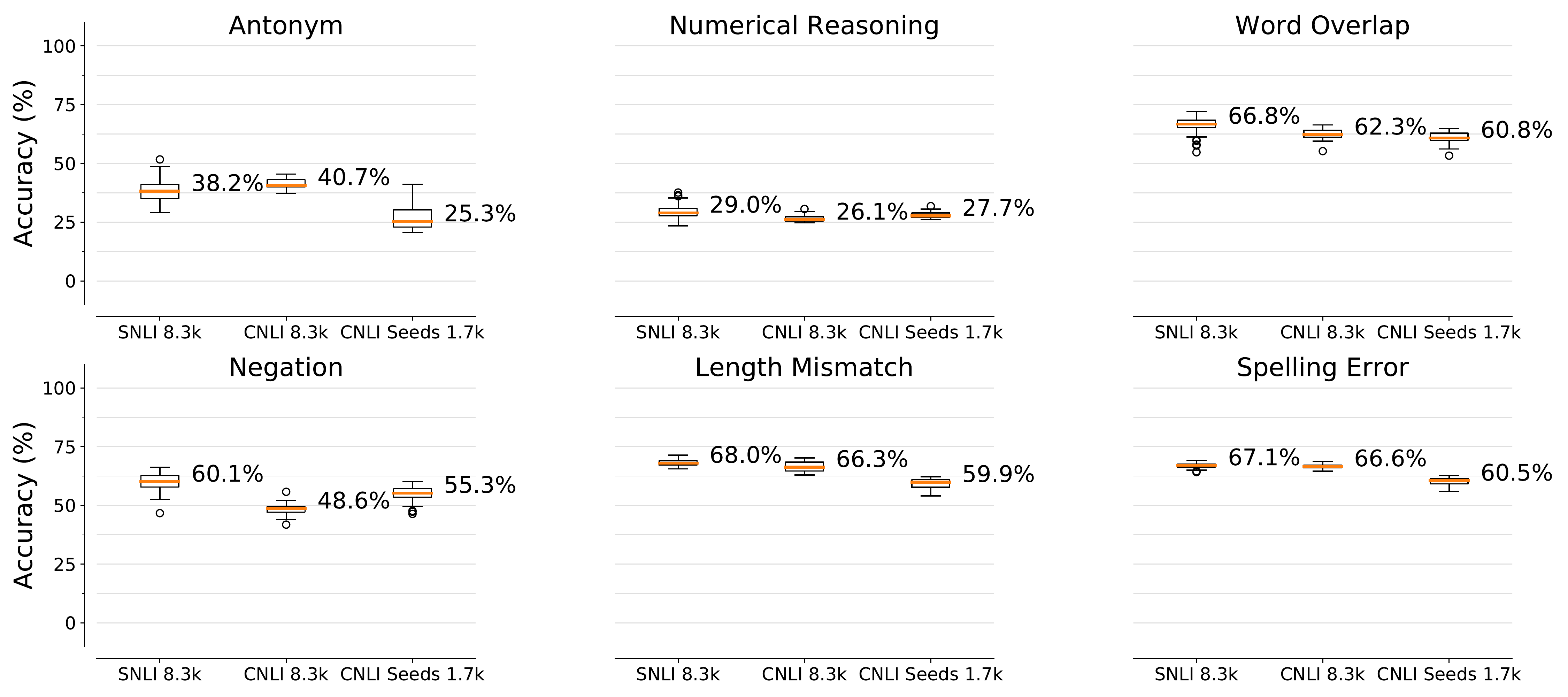}
            \label{subfig:stress}
        }\\
    \subfloat[b][]{
            \includegraphics[width=1\textwidth]{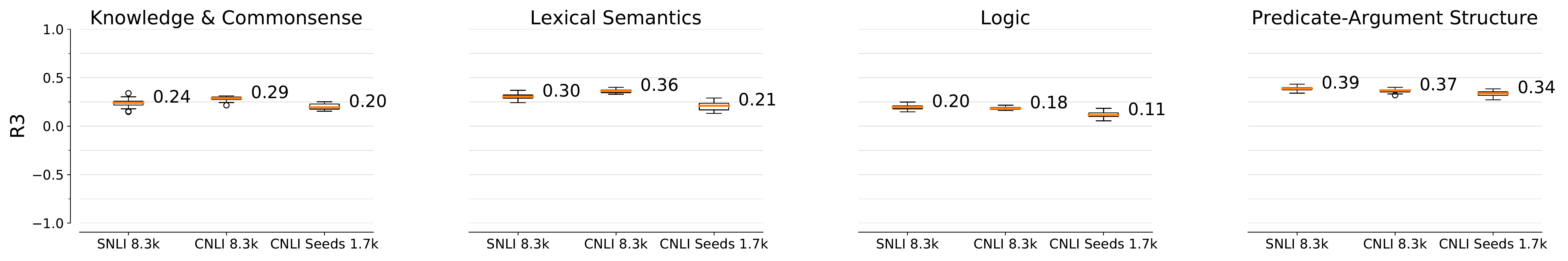}
            \label{subfig:GLUEdiagnostic}
        }
    \caption{Performance on diagnostic sets using (a) accuracy for \citet{naik2018NLIstress} examples and (b) $R_3$ score on GLUE diagnostic examples trained on subsampled SNLI, CNLI, and CNLI seed examples. Labels and orange lines indicate median scores.}
    \label{fig:challengesets}
\end{figure*}

Figure \ref{fig:challengesets} presents performances on the diagnostic sets from \citet{naik2018NLIstress} and \citet{wang2019GLUE}. For the GLUE diagnostic sets, we follow the authors and use $R_3$ \citep{gorodkin04R3} as our evaluation metric. The distributions of classification accuracy again show that CNLI yields similar performance compared to unaugmented datasets of similar size on most of the categories.

However, we find that training on CNLI yields worse performance than using either unaugmented SNLI or CNLI seed examples for Negation examples. These challenge examples append the phrase \textit{``and false is not true''} to every hypothesis in the MNLI validation set. This construction introduces the strong negation word \textit{``no''} to target the association between negation words and the \textit{contradiction} label without changing the truth condition of the sentence. We speculate that the augmented data may have amplified this association already present among the seed examples. Not only does this show that CNLI can yield models that are less robust to certain challenge examples, but it also provides evidence that adding substantial numbers of counterfactual examples to a dataset can hurt robustness. 

\paragraph{Lexical Diversity}

\begin{table}[t]
    \centering
    \resizebox{.45\textwidth}{!}{
    \begin{tabular}{l r r r}
        \toprule
        $n$ & \multicolumn{1}{l}{SNLI 8.3k} & \multicolumn{1}{l}{CNLI 8.3k} & \multicolumn{1}{l}{CNLI Seed 1.7k}\\
        \midrule
        1 & 6.2k & 4.8k  & 3.5k \\
        2 & 30.4k & 21.6k  & 13.0k \\
        3 & 52.3k & 36.3k  & 19.9k \\
        4 & 60.2k & 42.5k & 21.5k \\
        \bottomrule
    \end{tabular}}
    \caption{Number of unique $n$-gram types observed in each training set.}
    \label{tab:ngram_stats}
\end{table}

Given the minimal edits constraint in CNLI, we study the lexical diversity of the training sets to see the effectiveness of this constraint and whether the general improvement of CNLI over seed examples is a result of greater diversity from a larger training set. Table \ref{tab:ngram_stats} provides the number of n-grams present in each training set with $n$ varying from one to four. We see that including minimally edited examples to CNLI increases the number of $n$-grams present, roughly doubling the number of 4-grams, which corresponds to the general improvement over seed examples.

We also observe that CNLI contains roughly 70\% of 2-, 3-, and 4-grams compared to similarly large unaugmented training sets. This seems natural given the minimal edits constraint when collecting counterfactually-augmented examples and highlights the fact that this type of data augmentation results in less diversity per example.

\section{Conclusion}
\label{sec:Conclusion}

We follow a similar setup to \citet{khashabi2020pertboolq} and use English NLI data to test whether counterfactually-augmented training data yields models that generalize better to out-of-domain data and are more robust to challenge examples. We first find that adding counterfactually-augmented data improves generalization, but provides no advantage over adding similar amounts of unaugmented data. Further, we find that the improvement over seed examples corresponds to an increase in $n$-gram diversity. We also find that including counterfactually-augmented data can make models less robust to challenge examples. Assuming that crowdworkers take a similar amount of time to make targeted edits to examples and to write new examples \citep{Bowman2020EntailmentNewProtocols}, there is then no obvious value in crowdsourcing augmentations under current protocols for use as training data.

Despite these findings, we argue that there is still value in naturally collected counterfactually-augmented NLU data. \citet{gardner2020augmenteddata} show that collecting this type of data can be used as a method to address systematic gaps in testing data. As performances on benchmarks become saturated, we still view this style of augmenting test sets as a viable method to provide longer-lasting benchmarks in addition to standard test set creation.

The success of \citet{gardner2020augmenteddata} in using expert-designed counterfactual augmentation to target specific phenomena for \textit{evaluation} suggests that it may be possible to target heuristics in training data with expert guidance during the crowdsourcing process. Further, understanding how to identify heuristics to target and the types of useful augmentations to collect, assuming such a thing is possible, are important directions we leave to future work.

\section*{Acknowledgements}
We thank Clara Vania and Jason Phang for their helpful feedback and Alex Wang for providing the script for $n$-gram counts that we base our lexical diversity analysis code on. This project has benefited from financial support to SB by Eric and Wendy Schmidt (made by recommendation of the Schmidt Futures program), by Samsung Research (under the project \textit{Improving Deep Learning using Latent Structure}), by Intuit, Inc., and in-kind support by the NYU High-Performance Computing Center and by NVIDIA Corporation (with the donation of a Titan V GPU). This material is based upon work supported by the National Science Foundation under Grant No.  1922658. Any opinions, findings, and conclusions or recommendations expressed in this material are those of the author(s) and do not necessarily reflect the views of the National Science Foundation.

\bibliography{citations}

\begin{thebibliography}{17}
\expandafter\ifx\csname natexlab\endcsname\relax\def\natexlab#1{#1}\fi

\bibitem[{Bowman et~al.(2015)Bowman, Angeli, Potts, and
  Manning}]{bowman2015SNLI}
Samuel~R. Bowman, Gabor Angeli, Christopher Potts, and Christopher~D. Manning.
  2015.
\newblock \href {https://doi.org/10.18653/v1/d15-1075} {A large annotated
  corpus for learning natural language inference}.
\newblock In \emph{Proceedings of the 2015 Conference on Empirical Methods in
  Natural Language Processing, {EMNLP} 2015, Lisbon, Portugal, September 17-21,
  2015}, pages 632--642. The Association for Computational Linguistics.

\bibitem[{Bowman et~al.(2020)Bowman, Palomaki, Soares, and
  Pitler}]{Bowman2020EntailmentNewProtocols}
Samuel~R. Bowman, Jennimaria Palomaki, Livio~Baldini Soares, and Emily Pitler.
  2020.
\newblock Collecting entailment data for pretraining: New protocols and
  negative results.
\newblock In \emph{Proceedings of EMNLP}.

\bibitem[{Gardner et~al.(2020)Gardner, Artzi, Basmova, Berant, Bogin, Chen,
  Dasigi, Dua, Elazar, Gottumukkala, Gupta, Hajishirzi, Ilharco, Khashabi, Lin,
  Liu, Liu, Mulcaire, Ning, Singh, Smith, Subramanian, Tsarfaty, Wallace,
  Zhang, and Zhou}]{gardner2020augmenteddata}
Matt Gardner, Yoav Artzi, Victoria Basmova, Jonathan Berant, Ben Bogin, Sihao
  Chen, Pradeep Dasigi, Dheeru Dua, Yanai Elazar, Ananth Gottumukkala, Nitish
  Gupta, Hanna Hajishirzi, Gabriel Ilharco, Daniel Khashabi, Kevin Lin,
  Jiangming Liu, Nelson~F. Liu, Phoebe Mulcaire, Qiang Ning, Sameer Singh,
  Noah~A. Smith, Sanjay Subramanian, Reut Tsarfaty, Eric Wallace, Ally Zhang,
  and Ben Zhou. 2020.
\newblock \href {http://arxiv.org/abs/2004.02709} {Evaluating models' local
  decision boundaries via contrast sets}.

\bibitem[{Gorodkin(2004)}]{gorodkin04R3}
J.~Gorodkin. 2004.
\newblock \href
  {https://doi.org/https://doi.org/10.1016/j.compbiolchem.2004.09.006}
  {Comparing two k-category assignments by a k-category correlation
  coefficient}.
\newblock \emph{Computational Biology and Chemistry}, 28(5):367 -- 374.

\bibitem[{Gururangan et~al.(2018)Gururangan, Swayamdipta, Levy, Schwartz,
  Bowman, and Smith}]{gururangan2018artifactsNLI}
Suchin Gururangan, Swabha Swayamdipta, Omer Levy, Roy Schwartz, Samuel~R.
  Bowman, and Noah~A. Smith. 2018.
\newblock \href {https://doi.org/10.18653/v1/n18-2017} {Annotation artifacts in
  natural language inference data}.
\newblock In \emph{Proceedings of the 2018 Conference of the North American
  Chapter of the Association for Computational Linguistics: Human Language
  Technologies, NAACL-HLT, New Orleans, Louisiana, USA, June 1-6, 2018, Volume
  2 (Short Papers)}, pages 107--112. Association for Computational Linguistics.

\bibitem[{Kaushik et~al.(2020)Kaushik, Hovy, and Lipton}]{kaushik2020cnli}
Divyansh Kaushik, Eduard~H. Hovy, and Zachary~Chase Lipton. 2020.
\newblock \href {https://openreview.net/forum?id=Sklgs0NFvr} {Learning the
  difference that makes {A} difference with counterfactually-augmented data}.
\newblock In \emph{8th International Conference on Learning Representations,
  {ICLR} 2020, Addis Ababa, Ethiopia, April 26-30, 2020}. OpenReview.net.

\bibitem[{Khashabi et~al.(2020)Khashabi, Khot, and
  Sabharwal}]{khashabi2020pertboolq}
Daniel Khashabi, Tushar Khot, and Ashish Sabharwal. 2020.
\newblock More bang for your buck: Natural perturbation for robust question
  answering.
\newblock In \emph{Proceedings of EMNLP}.

\bibitem[{Liu et~al.(2019)Liu, Ott, Goyal, Du, Joshi, Chen, Levy, Lewis,
  Zettlemoyer, and Stoyanov}]{liu2019RoBERTa}
Yinhan Liu, Myle Ott, Naman Goyal, Jingfei Du, Mandar Joshi, Danqi Chen, Omer
  Levy, Mike Lewis, Luke Zettlemoyer, and Veselin Stoyanov. 2019.
\newblock \href {http://arxiv.org/abs/1907.11692} {{RoBERTa}: {A} robustly
  optimized {BERT} pretraining approach}.
\newblock \emph{CoRR}, abs/1907.11692.

\bibitem[{McCoy et~al.(2019)McCoy, Pavlick, and Linzen}]{mccoy2019HANS}
Tom McCoy, Ellie Pavlick, and Tal Linzen. 2019.
\newblock \href {https://doi.org/10.18653/v1/p19-1334} {Right for the wrong
  reasons: Diagnosing syntactic heuristics in natural language inference}.
\newblock In \emph{Proceedings of the 57th Conference of the Association for
  Computational Linguistics, {ACL} 2019, Florence, Italy, July 28- August 2,
  2019, Volume 1: Long Papers}, pages 3428--3448. Association for Computational
  Linguistics.

\bibitem[{Mosbach et~al.(2020)Mosbach, Andriushchenko, and
  Klakow}]{mosbach2020stabilityBERT}
Marius Mosbach, Maksym Andriushchenko, and Dietrich Klakow. 2020.
\newblock \href {http://arxiv.org/abs/2006.04884} {On the stability of
  fine-tuning {BERT:} misconceptions, explanations, and strong baselines}.
\newblock \emph{CoRR}, abs/2006.04884.

\bibitem[{Naik et~al.(2018)Naik, Ravichander, Sadeh, Ros{\'{e}}, and
  Neubig}]{naik2018NLIstress}
Aakanksha Naik, Abhilasha Ravichander, Norman~M. Sadeh, Carolyn~Penstein
  Ros{\'{e}}, and Graham Neubig. 2018.
\newblock \href {https://www.aclweb.org/anthology/C18-1198/} {Stress test
  evaluation for natural language inference}.
\newblock In \emph{Proceedings of the 27th International Conference on
  Computational Linguistics, {COLING} 2018, Santa Fe, New Mexico, USA, August
  20-26, 2018}, pages 2340--2353. Association for Computational Linguistics.

\bibitem[{Nie et~al.(2020)Nie, Williams, Dinan, Bansal, Weston, and
  Kiela}]{nie2020adversarialNLI}
Yixin Nie, Adina Williams, Emily Dinan, Mohit Bansal, Jason Weston, and Douwe
  Kiela. 2020.
\newblock \href {https://www.aclweb.org/anthology/2020.acl-main.441/}
  {Adversarial {NLI:} {A} new benchmark for natural language understanding}.
\newblock In \emph{Proceedings of the 58th Annual Meeting of the Association
  for Computational Linguistics, {ACL} 2020, Online, July 5-10, 2020}, pages
  4885--4901. Association for Computational Linguistics.

\bibitem[{Poliak et~al.(2018)Poliak, Naradowsky, Haldar, Rudinger, and
  Durme}]{poliak2018hypothesisonly}
Adam Poliak, Jason Naradowsky, Aparajita Haldar, Rachel Rudinger, and
  Benjamin~Van Durme. 2018.
\newblock \href {https://doi.org/10.18653/v1/s18-2023} {Hypothesis only
  baselines in natural language inference}.
\newblock In \emph{Proceedings of the Seventh Joint Conference on Lexical and
  Computational Semantics, *SEM@NAACL-HLT 2018, New Orleans, Louisiana, USA,
  June 5-6, 2018}, pages 180--191. Association for Computational Linguistics.

\bibitem[{Tsuchiya(2018)}]{tsuchiya18RTEhypothesis}
Masatoshi Tsuchiya. 2018.
\newblock \href
  {http://www.lrec-conf.org/proceedings/lrec2018/summaries/786.html}
  {Performance impact caused by hidden bias of training data for recognizing
  textual entailment}.
\newblock In \emph{Proceedings of the Eleventh International Conference on
  Language Resources and Evaluation, {LREC} 2018, Miyazaki, Japan, May 7-12,
  2018}. European Language Resources Association {(ELRA)}.

\bibitem[{Wang et~al.(2019{\natexlab{a}})Wang, Singh, Michael, Hill, Levy, and
  Bowman}]{wang2019GLUE}
Alex Wang, Amanpreet Singh, Julian Michael, Felix Hill, Omer Levy, and
  Samuel~R. Bowman. 2019{\natexlab{a}}.
\newblock \href {https://openreview.net/forum?id=rJ4km2R5t7} {{GLUE:} {A}
  multi-task benchmark and analysis platform for natural language
  understanding}.
\newblock In \emph{7th International Conference on Learning Representations,
  {ICLR} 2019, New Orleans, LA, USA, May 6-9, 2019}. OpenReview.net.

\bibitem[{Wang et~al.(2019{\natexlab{b}})Wang, Tenney, Pruksachatkun, Yeres,
  Phang, Liu, Htut, , Yu, Hula, Xia, Pappagari, Jin, McCoy, Patel, Huang,
  Grave, Kim, F\'evry, Chen, Nangia, Mohananey, Kann, Bordia, Patry, Benton,
  Pavlick, and Bowman}]{wang2019jiant}
Alex Wang, Ian~F. Tenney, Yada Pruksachatkun, Phil Yeres, Jason Phang, Haokun
  Liu, Phu~Mon Htut, , Katherin Yu, Jan Hula, Patrick Xia, Raghu Pappagari,
  Shuning Jin, R.~Thomas McCoy, Roma Patel, Yinghui Huang, Edouard Grave,
  Najoung Kim, Thibault F\'evry, Berlin Chen, Nikita Nangia, Anhad Mohananey,
  Katharina Kann, Shikha Bordia, Nicolas Patry, David Benton, Ellie Pavlick,
  and Samuel~R. Bowman. 2019{\natexlab{b}}.
\newblock \texttt{jiant} 1.3: A software toolkit for research on
  general-purpose text understanding models.
\newblock \url{http://jiant.info/}.

\bibitem[{Williams et~al.(2018)Williams, Nangia, and Bowman}]{williams2018MNLI}
Adina Williams, Nikita Nangia, and Samuel~R. Bowman. 2018.
\newblock \href {https://doi.org/10.18653/v1/n18-1101} {A broad-coverage
  challenge corpus for sentence understanding through inference}.
\newblock In \emph{Proceedings of the 2018 Conference of the North American
  Chapter of the Association for Computational Linguistics: Human Language
  Technologies, {NAACL-HLT} 2018, New Orleans, Louisiana, USA, June 1-6, 2018,
  Volume 1 (Long Papers)}, pages 1112--1122. Association for Computational
  Linguistics.

\end{thebibliography}
\bibliographystyle{acl_natbib}


\end{document}